\documentclass[conference]{IEEEtran}
\IEEEoverridecommandlockouts

\usepackage{cite}
\usepackage{amsmath,amssymb,amsfonts}
\usepackage{algorithmic}
\usepackage{graphicx}
\usepackage{textcomp}
\usepackage{comment}
\usepackage{xcolor}
\usepackage{dsfont}
\usepackage{booktabs}
\usepackage{makecell}

\def\BibTeX{{\rm B\kern-.05em{\sc i\kern-.025em b}\kern-.08em
    T\kern-.1667em\lower.7ex\hbox{E}\kern-.125emX}}
\begin{document}

\title{Insights into the Unknown: Federated Data Diversity Analysis on Molecular Data}

\author{\IEEEauthorblockN{Markus Bujotzek}
\IEEEauthorblockA{\textit{Apheris AI} \\
Berlin, Germany }
\IEEEauthorblockA{\textit{MIC at German Cancer Research Center} \\
Heidelberg, Germany}
\IEEEauthorblockA{\textit{Medical Faculty of Heidelberg University} \\
Heidelberg, Germany}
\texttt{m.bujotzek@apheris.com}
\and
\IEEEauthorblockN{Evelyn Trautmann}
\IEEEauthorblockA{\textit{Apheris AI} \\
Berlin, Germany}
\and
\IEEEauthorblockN{Calum Hand}
\IEEEauthorblockA{\textit{Apheris AI}\\
Berlin, Germany} \\
\and
\IEEEauthorblockN{Ian Hales}
\IEEEauthorblockA{\textit{Apheris AI}\\
Berlin, Germany} \\
}

\maketitle

\begin{abstract}

AI methods are increasingly shaping pharmaceutical drug discovery.
However, their translation to industrial applications remains limited due to their reliance on public datasets, lacking scale and diversity of proprietary pharmaceutical data.
Federated learning (FL) offers a promising approach to integrate private data into privacy-preserving, collaborative model training across data silos.
This federated data access complicates important data-centric tasks such as estimating dataset diversity, performing informed data splits, and understanding the structure of the combined chemical space.
To address this gap, we investigate how well federated clustering methods can disentangle and represent distributed molecular data.
We benchmark three approaches, Federated kMeans (Fed-kMeans), Federated Principal Component Analysis combined with Fed-kMeans (Fed-PCA+Fed-kMeans), and Federated Locality-Sensitive Hashing (Fed-LSH), against their centralized counterparts on eight diverse molecular datasets.
Our evaluation utilizes both, standard mathematical and a chemistry-informed evaluation metrics, \textit{SF-ICF}, that we introduce in this work.
The large-scale benchmarking combined with an in-depth explainability analysis shows the importance of incorporating domain knowledge through chemistry-informed metrics, and on-client explainability analyses for federated diversity analysis on molecular data.

\end{abstract}


\section{Introduction}
AI methods are increasingly transforming pharmaceutical drug discovery research, with breakthroughs such as Alphafold for 3D protein structure prediction \cite{abramson2024accurate}, and foundation models for molecular property prediction \cite{ross2022large}.
While these models show impressive capabilities, their performance in industrial settings remains limited, primarily because they are trained on public datasets that lack the scale and relevance of proprietary pharmaceutical data \cite{callaway2025alphafold}.
Integrating this private data into model training is the critical next step toward realizing the effective impact of AI in drug discovery.

The paradigm of federated learning (FL) offers an approach to unlock this potential by enabling privacy-preserving, collaborative model training across organizational boundaries, with demonstrated success in drug discovery applications \cite{heyndrickx2023melloddy}.
While FL addresses the issue of data scarcity by making proprietary data accessible for model training, it introduces a major hurdle: the lack of global insight into the data's properties and diversity between distributed parties. 
The distributed data access makes it inherently difficult to construct meaningful training, validation, and test splits, a critical requirement for avoiding overoptimistic performance estimates \cite{simm2021splitting}. Moreover, essential tasks such as data valuation and the identification of overlapping or sparse regions in the distributed data space become non-trivial in federated settings, thereby motivating the need for privacy-preserving methods for federated data diversity analysis.

Approaches such as clustering, dimensionality reduction, and data visualization are commonly employed to gain detailed insights into high-dimensional chemical datasets containing up to millions of molecular structures.
In centralized settings, clustering finds a wide application in semantically disentangling high-dimensional molecular representations.
Researchers have compared general-purpose clustering algorithms with chemistry-specific techniques like Taylor–Butina clustering \cite{hernandez2023best}, explored deep learning-based approaches such as clustering based on Variational Autoencoder (VAE) \cite{hadipour2022deep}, and developed tools like ChemMine \cite{backman2011chemmine} for interactive analysis and clustering of small molecules.
Similarly, visualization and dimensionality reduction techniques are widely used to better understand the diversity and structure of chemical datasets. In addition to general-purpose methods such as Principal Component Analysis (PCA), Uniform Manifold Approximation and Projection (UMAP) \cite{mcinnes2018umap}, and t-distributed Stochastic Neighbor Embedding (t-SNE) \cite{van2008visualizing}, chemistry-specific tools like Tree MAP (TMAP) \cite{probst2020visualization} have gained popularity for their ability to reveal and visualize structural relationships and diversity within large chemical datasets \cite{orlov2025high}.
However, the applicability of these techniques is largely limited to centralized settings, where pairwise similarities between molecular data points can be directly computed \cite{qiao2025federated}.

The translation of the aforementioned approaches to operate on distributed data in federated settings is challenging due to lack of global data access and visibility.
Nevertheless, few federated clustering methods have successfully extended the intuitive k-Means clustering \cite{kumar2020federated, garst2024federated} and spectral clustering \cite{qiao2023federated, qiao2025federated} to decentralized settings. Similarly, dimensionality reduction techniques have been adapted to the federated paradigm, including Federated PCA \cite{grammenos2020federated}, as well as federated versions of UMAP and t-SNE \cite{qiao2025federated}.

The application of federated data analysis techniques to distributed molecular data is sparse.
Simm et al. \cite{simm2021splitting} investigated the specific challenge of generating informed training, validation, and test splits across federated clients to prevent structurally similar molecules from appearing in both training and test sets; an issue that can lead to overly optimistic performance estimates and poor real-world generalization.
Thereby, Simm et al. compare several federated clustering methods based on locality-sensitive hashing (LSH) \cite{gionis1999similarity}, Taylor-Butina clustering \cite{butina1999unsupervised}, and scaffold-based binning, against randomly assigned folds, as they have been found to be well-performing in centralized scenarios.
While these methods are highly relevant to data diversity analysis, their evaluation is narrowly focused on data splitting. 
As such, the broader question of which clustering approach most effectively disentangles and represents the structure of distributed chemical data remains open.

Motivated by this gap in the interdisciplinary literature, we pose the following research question:
\textit{How can federated clustering methods most meaningful disentangle and cluster molecular data in distributed settings?}
We address this research question through the following key contributions:
\begin{itemize}
    \item We benchmark federated k-Means (Fed-kMeans) clustering, a federated PCA dimension reduction combined with a Fed-kMeans (Fed-PCA+Fed-kMeans), and federated LSH (Fed-LSH) against their centralized counterparts.
    \item Besides standard mathematical cluster evaluation metrics, we propose the chemistry-informed Scaffold-Frequency Inverse-Cluster-Frequency (SF-ICF) evaluation metric.
    \item The methods are evaluated on 8 large and diverse molecular datasets.
    \item Clustering performance is quantitatively evaluated using both mathematical and chemistry-informed evaluation metrics, and further enriched by an in-depth clustering explainability analysis.
\end{itemize}

\section{Materials and Methods}
\subsection{Data}
We benchmark the proposed federated clustering methods using the PharmaBench data collection \cite{niu2024pharmabench}, which provides not only well-curated molecular structures and associated properties, but also rich metadata from the ChEMBL database.
This metadata is representative for real-world pharmaceutical applications and serves as a valuable resource for the explainability of clustering outcomes in both research and deployment scenarios.
The raw PharmaBench dataset comprises eight subsets, containing 6,051; 14,107; 16,822; 14,775; 25,332; 3,381; 39,665; and 900 molecules, respectively.
Since our focus lies on unsupervised clustering, we discard the provided classification and regression ground truth.
Instead, we process and enrich the dataset by computing Murcko scaffolds \cite{schuffenhauer2007scaffold}, which abstract the molecular structure by retaining core ring systems and linkers while removing side chains of the molecule.
Additionally, we derive extended-connectivity fingerprints \cite{rogers2010extended} (ECFPs; radius = 2, 2048 bits) using the RDKit toolkit \cite{landrum2013rdkit}, yielding high-dimensional binary vectors encoding the presence or absence of local atomic environments.
The ECFP provide structurally informative inputs for the clustering algorithms, and the scaffolds serve as the basis for chemistry-informed evaluation metrics.

\subsection{Federated Clustering Methods}
In our experimental benchmarking, we compare Fed-kMeans, Fed-PCA combined with Fed-kMeans, and Fed-LSH against their centralized counterparts, which serve as upper baselines, as well as against a random clustering assignment representing a lower, noise-based baseline.
Prior to the federated evaluation, we conducted an extensive comparison of various clustering algorithms in a centralized setting using the same data and same evaluation metrics. Despite its simplicity, k-Means combined with a preceding PCA emerged as the most effective centralized approach, further motivating its adaptation to the federated setting.

\subsubsection{Federated k-Means Clustering}
For the Fed-kMeans clustering, we adapted the NVIDIA FLARE federated k-means example to operate on molecular data. The implementation is based on the mini-batch k-means algorithm, with the distinction that data is not partitioned into mini-batches but rather distributed across FL clients.
In each communication round, FL clients perform local k-means clustering on their respective datasets using the current global centroids, or k-means++ \cite{arthur2006k} initialization in the first round. Each client then sends its locally updated centroids and corresponding cluster counts to the server, which computes a weighted average (based on cluster sizes) to update the global centroids. These updated centroids are subsequently communicated back to the FL clients for the next iteration.

\subsubsection{Federated Principal Component Analysis}
Our federated PCA implementation is designed for a horizontal FL setting, meaning that all FL clients $N$ with local data $X^n_i, i=1,...,D_n$ share the same feature space $R^F$.
To compute the PCA in a distributed manner, we first need to construct the joint covariance matrix $\mathcal{C}$ across all FL clients $N$ on their local data $X^n_i$.
\begin{align}
    \mathcal{C}(f_y, f_z) = \sum_{n \in N} \sum_{i=1}^{D_n} 
    &\big(X^n_i(f_y) - \overline{X(f_y)}\big)\big(X^n_i(f_z) - \overline{X(f_z)}\big)
    \label{eq:cov}
\end{align}

Equation \ref{eq:cov} is fully decomposable across the FL clients $N$ and can thus be computed exactly, without introducing any error due to federation.
Once the global covariance matrix $\mathcal{C}$ is computed, our method follows the standard PCA computation by computing centrally the eigenvalue decomposition and selecting the $p$ dominant eigenvalues, yielding the final Fed-PCA projection matrix $\mathcal{P}$.
Subsequently, this matrix $\mathcal{P}$ is distributed to the FL clients, where the local data dimensionality is reduced from $F$ to $p$ dimensions.

\subsubsection{Federated Locality-sensitive Hashing}
LSH relies on a binary molecular representation, such as ECFPs, and aims to group structurally similar molecules by exploiting high-entropy fingerprint bits.
The core idea is to identify the $N_{he}$ bits with the highest entropy across a set of molecular fingerprints, as these bits carry the most discriminative information for distinguishing molecular structures. Molecules that share identical values for these $N_{he}$ high-entropy bits are grouped into the same bin, resulting in up to $2^{N_{he}}$ bins for binary fingerprints. We follow the implementation provided by Simm et al. \cite{simm2021splitting}.

To extend LSH to a federated setting, the identification of high-entropy bits must be performed collaboratively. Each client locally computes its top $N_{he,n}$ high-entropy bit positions and transmits the corresponding indices to the FL server. Since fingerprinting is applied consistently across all clients, these bit indices are semantically aligned across the federation. The server then computes the intersection of the locally reported high-entropy bit sets, yielding a consensus set of $N_{he,C}$ global high-entropy bits.
This shared set is distributed back to the clients, which then perform clustering by grouping molecules that share identical values across the $N_{he,C}$ fingerprint bits.
Should the intersection set $N_{he,C}$ be empty, a larger number $N_{he,n}$ of local high-entropy bits has to be selected.

\subsection{Evaluation metrics}
We differentiate our evaluation metrics in two categories, mathematical and chemistry-informed metrics.
To evaluate the general clustering quality we employ mathematical evaluation metrics: The Silhouette Score \cite{rousseeuw1987silhouettes} measures how well samples are clustered based on intra-cluster cohesion and inter-cluster separation; the Calinski-Harabasz (CH) score \cite{calinski1974dendrite} captures the ratio of inter-cluster to intra-cluster variance; and the Davies-Bouldin (DB) score \cite{davies2009cluster} assesses cluster similarity with lower values indicating better separation and compactness. All mathematical metrics are computed using Euclidean distance and are implemented via the scikit-learn library.

Beyond mathematical evaluation metrics, we also evaluate the clustering using chemistry-informed metrics.
Acknowledging that Euclidean distance becomes with increasing dimensionality less meaningful \cite{reutlinger2012nonlinear}, the Tanimoto distance is a widely adopted distance metric, corresponding to the Jaccard or Dice index.
We capture the inter- and intra-cluster Tanimoto distances by computing the Tanimoto distance matrices between all molecule's fingerprints of a FL client, and using these pre-computed distance matrices in a tanimoto-based silhouette score.
Moreover, we capture how clustering correlates with binning molecules based on their scaffolds using a metric inspired by the term-frequency inverse-document-frequency (TF-IDF) metric from the domain of text-based information retrieval.
We compute for a scaffold $s_i$ in a cluster $c_j \in C$ the scaffold frequency $SF(s_i, c_j)$(\ref{eq:sf}), and the inverse cluster frequency $ICF(s_i, C)$ (\ref{eq:icf}), i.e. the intra-cluster purity ($SF$) and the inter-cluster rarity ($ICF$) of scaffolds.
Subsequently, we compute the weighted sums obtaining one final \textit{SF-ICF} score for all clusters and scaffolds (\ref{eq:SF-ICF}). 
\begin{equation}
    SF(s_i, c_j) = \frac{|\{s_i \in c_j\}|}{|c_j|}
    \label{eq:sf}
\end{equation}
\begin{equation}
    ICF(s_i, C) = \frac{\log\left(\frac{|C|}{\sum_j\mathds{1}_{s_i \in  c_j}}\right)}{\log(|C|)}
    \label{eq:icf}
\end{equation}
\begin{equation}
    SF-ICF = \sum_{j=1}^{|C|} \frac{|c_j|}{N} \left( \sum_{s_i \in c_j} SF(s_i, c_j) \cdot ICF(s_i, C) \right)
    \label{eq:SF-ICF}
\end{equation}

Note that the \textit{SF-ICF} score can be generalized capturing the fragmentation across the clusters with respect to (w.r.t.) any other meta information which we use for explainability of the clustering results.

\begin{figure*}[t]
  \centering
  \includegraphics[width=\linewidth]{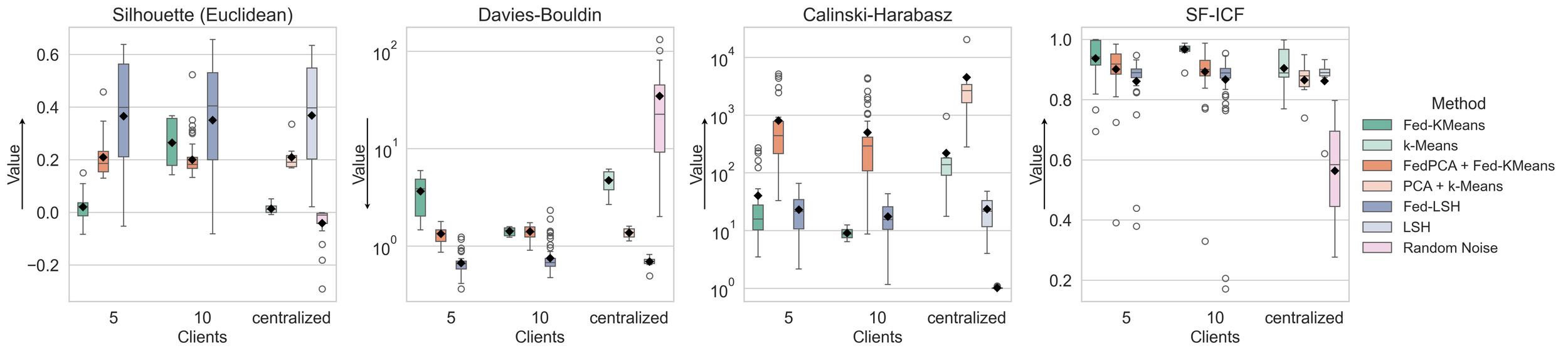}
    \caption{Benchmarking results of federated clustering (Fed-kMeans, Fed-PCA+Fed-kMeans, Fed-LSH) methods against their centralized counterparts. The methods are compared w.r.t. 3 mathematical (Silhouette (Euclidean), Davies-Bouldin, Calinski-Harabasz), and the chemistry-informed evaluation metric (SF-ICF). Mean values are marked with a diamond, arrows point towards better metric values.}
    \label{fig:flta-results}
\end{figure*}

\section{Experiments}
For our benchmarking experiments, each of the eight PharmaBench datasets was partitioned across FL clients based on molecular scaffolds. First, we computed the Murcko scaffolds for all molecules in a given dataset and then assigned an approximately equal number of unique scaffolds to each FL client. We varied the number of clients between 5 and 10 as a hyperparameter.
To better reflect realistic deployment scenarios, where it is unlikely that molecular scaffolds are entirely isolated to a single pharmaceutical company, we adopted a soft scaffold assignment strategy.
For each scaffold, approximately 90\% of its associated molecules (sampled via a Dirichlet distribution) were allocated to a primary client, while the remaining 10\% were randomly distributed among the other clients.
This splitting scheme results in a uniform distribution of scaffolds across clients, while mimicing scaffold overlap and yielding realistic variations in the number of molecules per client, all within the same order of magnitude.

Besides investigating the federated clustering methods across different numbers of FL clients (5, 10), we vary the following hyperparameters per method:
For Fed-kMeans, we vary the number of $k$ clusters between 5, 10, 20, 50, 100, 200, 500 and the number of FL rounds $r$ between 3, 5, 10.
In combination with Fed-kMeans and its hyperparameters, we vary for the Fed-PCA the number of principal components $p$ between 5, 10, 20, 50.
For Fed-LSH, we vary the locally identified $N_{he}$ highest-entropy bits between 4, 8, 16, 32.

Prior to benchmarking, we determine the optimal hyperparameter configuration for each federated clustering method through extensive hyperparameter tuning.
These optimized configurations are then used to benchmark the methods against one another, as well as against centralized upper and lower baselines.
The lower baseline randomly assigns molecules to clusters and serves as a noise reference, helping to contextualize whether a clustering method achieves meaningful structure beyond randomness.
The upper baseline corresponds to the centralized version of each clustering method using its optimal hyperparameters demonstrating performances achievable if all data were centrally available.

\section{Results and Discussion}

\subsection{Hyperparameter Grid Search}
We performed a dense grid search to identify optimal hyperparameter configurations for each method.
Hyperparameters were selected based on aggregated rankings across FL client and datasets to ensure generalizability.
Methods were ranked using both mathematical and chemistry-informed metrics; these ranks were averaged across FL clients and then across all datasets to determine the best overall configuration per method.
The final optimal hyperparameters per method are:
Fed-kMeans with $k=5$ and $r=3$ for 5 FL clients and $k=500$ and $r=10$ for 10 FL clients. Fed-PCA + Fed-kMeans with $p=5$, $k=5$ and $r=3$ for 5 FL clients and $p=5$, $k=5$ and $r=5$ for 10 FL clients. Fed-LSH with $N_{he}=32$ for 5 and 10 FL clients.
The subsequent benchmarking only compares experiments computed using these optimal hyperparameters.

\subsection{Benchmarking Results}

The benchmarking results of the compared federated clustering methods are presented in figure \ref{fig:flta-results} evaluated across mathematical and chemistry-informed metrics.

Regarding the mathematical metrics (Silhouette (Euclidean), DB and CH scores), Fed-LSH achieves the best scores across both FL client settings for the Euclidean-based Silhouette and DB scores, indicating geometrically strong intra-cluster cohesion and inter-cluster separation.
In contrast, Fed-PCA+Fed-kMeans demonstrates best performance on the CH score, showing balanced clusters w.r.t. the ratio of inter- and intra-cluster variances.
For the chemistry-informed metrics, Fed-kMeans and Fed-PCA+Fed-kMeans achieve the best scores on the SF-ICF metric, indicating that the resulting clusters align well with underlying molecular scaffold structures.  
All centralized and federated perform poorly on the tanimoto-based Silhouette score, with values around or below 0.0, and in some cases worse than the random noise baseline. Due to its lack of discriminative power in this context, we exclude the tanimoto-based Silhouette score from further results and discussion.

Comparing the federated clustering methods to their centralized counterparts, the performance gap is smaller than expected, with federated approaches in some cases even outperforming the centralized baselines across certain metrics.
It is important to note that the comparison for SF-ICF is biased due to the scaffold-based data splitting across FL clients. This setup provides the federated methods with an implicit advantage by pre-grouping molecules with similar scaffolds on FL clients, resulting in fewer unique scaffolds easing better SF-ICF scores.
Analyzing the clustering results in context of the random noise clustering, we observe that all methods constantly outperform the noise baseline across all metrics (except for the tanimoto-based Silhouette score as mentioned and excluded).

Considering the best performing hyperparameter configurations selected for each clustering method across different numbers of FL clients, we observe that Fed-kMeans is highly sensitive to the selection of $k$.
While larger $k$ values improve performance on the Euclidean-based Silhouette and SF-ICF scores, they lead to performance drops on other metrics, potentially due to overclustering of the data.
The hyperparameters for Fed-PCA+Fed-kMeans for 5 and 10 FL clients only differ by the numbers of FL rounds $r$.
The method seems to be robust to a variation of FL clients, 
performing with an exception of CH score across both FL experiments close or even better than the centralized PCA+kMeans.
Also LSH and Fed-LSH over both FL client settings use the same hyperparameters of $N_{he}=32$, showing comparable performances and therefore robustness w.r.t. number of FL clients across both federated and centralized settings.

Overall, the quantitative results indicate that Fed-LSH performs best according to standard mathematical clustering metrics.
However, incorporating the chemistry-informed SF-ICF score provides valuable insight into the chemical relevance of the clusters, revealing that both Fed-kMeans variants form chemically meaningful clusters.

\subsection{Explainability}
Our benchmarking evaluates federated clustering methods using both mathematical and chemistry-informed metrics.
However, as different methods perform best under different metrics, the quantitative results remain ambiguous.
To address this uncertainty, we incorporate explainability to gain deeper insight into the clustering behavior.
Specifically, we analyze the clustering results of each method on the PharmaBench \textit{ames} dataset for FL client 1 of 5, focusing on feature group importance as well as cluster counts and size statistics to provide an exemplary interpretation of the formed clusters.

\subsubsection{Feature Group Importance}
To explain the formed clusters, we reverse-engineer them by analyzing feature group importance to identify metadata feature groups within the FL client's dataset that most strongly correlate with the cluster assignments and are thus considered \textit{important}.
To this end, we train a Random Forest classifier using the available categorical and numerical metadata features to predict the assigned cluster labels.
We then extract the feature importances from the trained model and aggregate them by feature group, resulting in a ranked list of feature group importances.

\begin{figure}
    \centering
    \includegraphics[width=\linewidth]{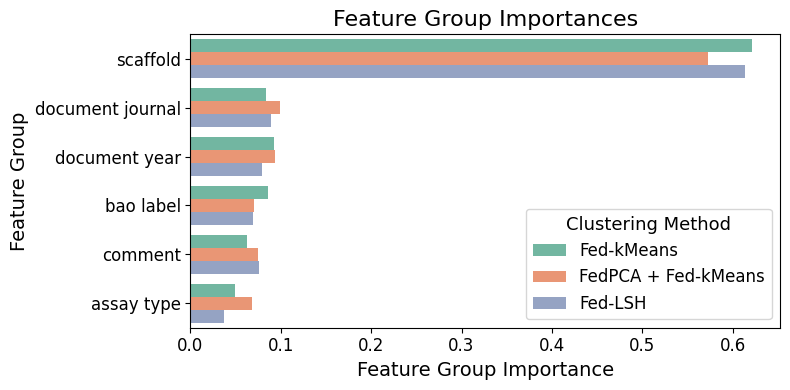}
    \caption{Feature group importances of meta information for federated clustering methods of FL client 1 of 5 on \textit{ames} data.}
    \label{fig:feature_group_importance}
\end{figure}

The feature group importances of the compared federated clustering methods are visualized in figure \ref{fig:feature_group_importance}.
Across all methods, the most important feature group are the \textit{scaffold} structures of clustered molecules, which aligns with high SF-ICF scores in the quantitative benchmarking and underlines the predictive power of the metric.
Not only the feature group importance of the scaffolds can be quantitatively supported via the SF-ICF score metric, the \textit{X}-F-ICF score can be generalized for all other feature groups, see table \ref{tab:xf_icf_scores}.
The \textit{X}-F-ICF scores indicate similar feature group importance as identified via the Random Forest based analysis, indicating consistency between qualitative explainability and quantitative metric-based evaluations.

\begin{table}[t]
    \centering
    \begin{tabular}{lccc}
        \toprule
        \makecell{\textbf{\{Feature Group\}} \\ \textbf{-F-ICF}} & \textbf{Fed-kMeans} & \textbf{\makecell{Fed-PCA \\ + Fed-kMeans}} & \textbf{Fed-LSH} \\
        \midrule
        Scaffold-F-ICF & 0.997 & 0.927 & 0.875 \\
        Document-Journal-F-ICF & 0.424 & 0.233 & 0.308 \\
        Document-Year-F-ICF & 0.642 & 0.383 & 0.518 \\
        Bao-Label-F-ICF & 0.333 & 0.124 & 0.226 \\
        Comment-F-ICF & 0.392 & 0.216 & 0.279 \\
        Assay-Type-F-ICF & 0.245 & 0.165 & 0.185 \\
        \bottomrule
    \end{tabular}
    \caption{Mean \{Feature Group\}-F-ICF scores for most important feature groups across federated clustering methods of FL client 1 of 5 on \textit{ames} data.}
    \label{tab:xf_icf_scores}
\end{table}

\subsubsection{Cluster Counts and Sizes}
To further interpret the clustering behavior, we examine the number of clusters and cluster size distributions (table \ref{tab:cluster_statistics}), both properties not directly captured by our benchmarking evaluation metrics.
As defined by the hyperparameter $k$ both, Fed-kMeans and Fed-PCA+Fed-kMeans produce 5 clusters with varying sizes.
In contrast, Fed-LSH generates a significantly larger number of clusters (480), below the theoretical maximum of $2^{N_{he}}$ clusters, resulting in a much finegrained partitioning of the data.

\begin{table}[t]
    \centering
    \begin{tabular}{lcccc}
        \toprule
        \textbf{Method} & \textbf{\# Clusters} & \textbf{Min Size} & \textbf{Max Size} & \textbf{Mean Size} \\
        \midrule
        Fed-kMeans & 5 & 1 & 1028 & 237.2 \\
        \makecell[l]{Fed-PCA+ \\ Fed-kMeans} & 5 & 43 & 819 & 237.2 \\
        Fed-LSH & 480 & 1 & 54 & 2.47 \\
        \bottomrule
    \end{tabular}
    \caption{Cluster count and size statistics of federated clustering methods of FL client 1 of 5 on \textit{ames} data.}
    \label{tab:cluster_statistics}
\end{table}

\begin{table}[ht]
    \centering
    \begin{tabular}{lcccc}
        \toprule
        \textbf{Feature Group} & \textbf{Unique Features} & \textbf{Mean \#} & \textbf{Min \#} & \textbf{Max \#}\\
        \midrule
        Scaffold & 331 & 3.58 & 1 & 71 \\
        \makecell[l]{Document- \\ Journal} & 12 & 89.42 & 1 & 363 \\
        \makecell[l]{Document- \\ Year} & 26 & 41.31 & 1 & 161 \\
        Bao-Label & 9 & 131.78 & 5 & 527 \\
        Comment & 10 & 118.60 & 1 & 287 \\
        Assay Type & 5 & 237.20 & 61 & 695 \\
        \bottomrule
    \end{tabular}
    \caption{Statistics of molecules sharing the most important features of FL client 1 of 5 on \textit{ames} data. }
    \label{tab:feature_stats}
\end{table}

To better understand the observed differences in cluster counts and sizes, we revisit the most important feature groups.
Table \ref{tab:feature_stats} presents summary statistics for these feature groups on FL client's dataset, which contains 1,186 molecules.
Notably, the top-ranked feature group, \textit{scaffold}, partitions the data most finely, with an average of 3.58 molecules sharing the same scaffold.
In contrast, other important feature groups aggregate molecules more coarsely, resulting in fewer, larger groups.

\begin{figure}
    \centering
    \includegraphics[width=\linewidth]{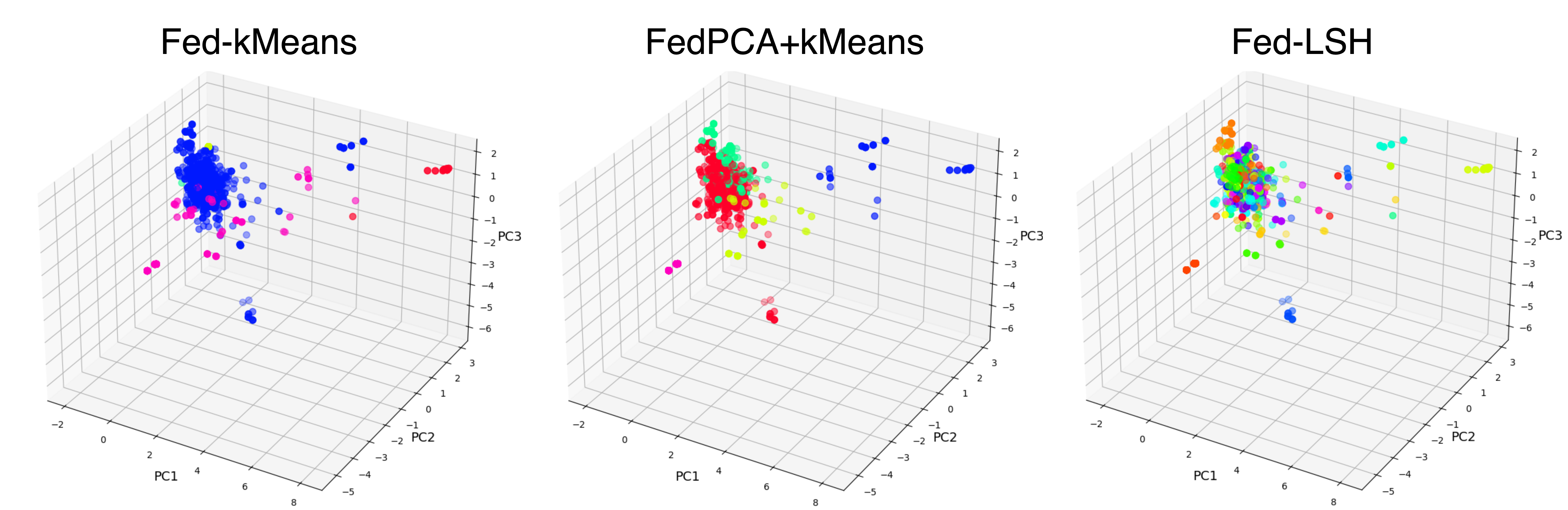}
    \caption{Qualitative clustering results of federated clustering methods on FL client 1 of 5 on \textit{ames} data. The high-dimensional data was dimension-reduced by a PCA with $p=3$.}
    \label{fig:qual_clustering}
\end{figure}

By cross-referencing tables \ref{tab:cluster_statistics} and \ref{tab:feature_stats}, we observe that Fed-LSH tends to overcluster the data: its mean cluster size is smaller than the average number of molecules sharing the same scaffold, the most important feature group.
In other words, even clustering the molecules solely based on shared scaffolds would result in a coarser partitioning than that produced by Fed-LSH.
The overclustering of Fed-LSH is illustrated in the qualitative clustering results in figure \ref{fig:qual_clustering}.

\subsection{SF–ICF and its role in molecular clustering evaluation}
\begin{figure}
    \centering
    \includegraphics[width=\linewidth]{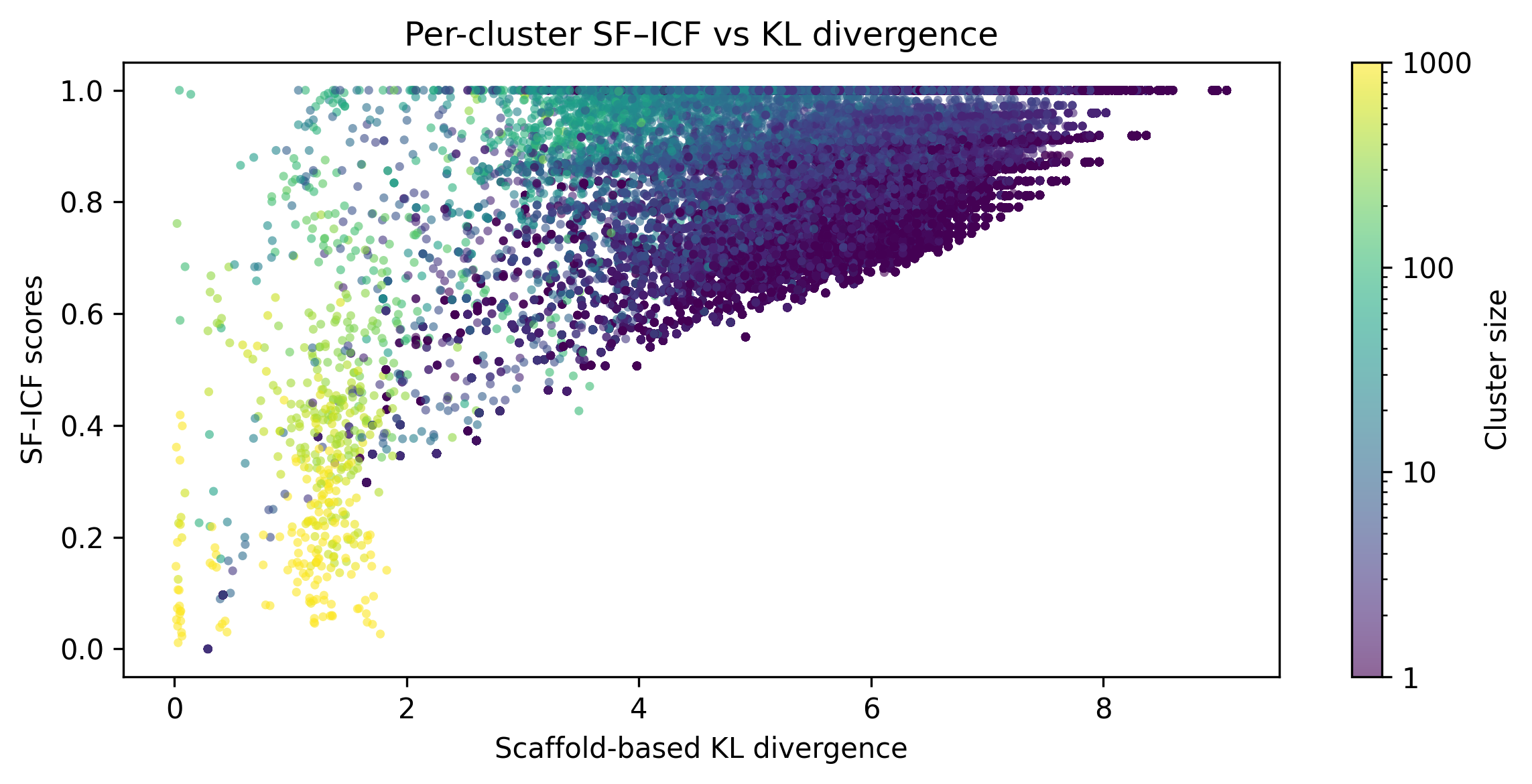}
    \caption{Per-cluster SF–ICF versus scaffold-based KL divergence over all federated experiments. Each point represents a cluster; color encodes cluster size.}
    \label{fig:sficf_vs_kld}
\end{figure}

We introduced SF-ICF as a chemistry-informed metric combining intra-cluster purity with inter-cluster rarity of clustered molecules' scaffolds.
To position it to established metrics, we compare per-cluster SF-ICF scores to the scaffold-based Kullback-Leibler divergence (KLD) $D_{KL}(P(S|c)||P(S))$ under consideration cluster sizes, see figure \ref{fig:sficf_vs_kld}.
KLD quantifies the divergence between cluster’s scaffold distribution and the global (dataset-wide) scaffold distribution.
We observe agreement at the extremes: small, scaffold-focused clusters score high on both metrics, while large, heterogeneous clusters score low on both.
However, we observe a deviation from their correlation for mid-sized clusters with high SF-ICF scores and lower KLD scores.
This deviation is an expected behavior of SF-ICF as it is designed to account for scaffold intra-cluster purity and inter-cluster fragmentation, still rewarding clustering of similar molecules with different scaffolds while fragmentation across other clusters is low.
As a result, the scaffold distribution within clusters can resemble the overall distribution, even if certain scaffolds occur exclusively in a single cluster, i.e. high SF-ICF but low KLD scores.
Overall, SF-ICF complements standard mathematical clustering metrics and distribution divergence measures with critical domain knowledge, and provides with its generalization to any meta information valuable insights and explainability of clustering results.

\section{Conclusion}
This work presents a comparative study of three federated clustering methods, Fed-kMeans, Fed-PCA+Fed-kMeans and Fed-LSH.
We benchmarked their performance quantitatively using both standard mathematical and chemistry-informed evaluation metrics across two federated settings and their centralized counterpart.
However, the ambiguity observed in the quantitative results demonstrates that conventional clustering metrics alone are insufficient to assess clustering quality in the domain of molecular chemistry.
Our findings underscore the importance of complementing a global federated quantitative evaluation with domain-specific metrics and a local, on-client explainability-driven analysis.
We recommend that federated clusterings on molecular data should integrate chemistry-informed metrics such as the SF-ICF score proposed here, alongside local sanity checks.

Future work should include expanding the study to a broader range of federated clustering algorithms and adopting adaptive hyperparameter strategies. Benchmarking should span more datasets and more FL setups (data partitioning schemes across FL clients, FL client counts), as well as include statistical analysis of the results.
Regarding evaluation metrics, future research should provide a deeper mathematical analysis of the introduced SF-ICF score, and introduce a metrics accounting for cluster counts and sizes to better detect overclustering.

Our work takes a first step towards federated diversity analysis, enabling structured insight into distributed molecular data without compromising privacy.
By introducing a domain-aware metric and interpretable evaluation, it establishes a foundation for more robust and trustworthy federated learning in AI-driven drug discovery.



\bibliographystyle{IEEEtran}
\bibliography{references}

\end{document}